\documentclass[journal, 10pt, twocolumn, a4paper]{IEEEtran}
\usepackage{float}
\usepackage{subcaption} 

\usepackage{graphicx}   
\usepackage{amsmath}    
\usepackage{amssymb}    

\usepackage{booktabs}   
\usepackage{cite}       
\usepackage[table]{xcolor} 
\usepackage{caption} 
\definecolor{ieeehighlight}{rgb}{0.0,0.4,0.7}
\usepackage[pagebackref,breaklinks,colorlinks,allcolors=ieeehighlight]{hyperref}

\title{PG-ControlNet: A Physics-Guided ControlNet for Generative Spatially Varying Image Deblurring}

\title{PG-ControlNet: A Physics-Guided ControlNet for Generative Spatially Varying Image Deblurring}

\author{
\IEEEauthorblockN{Hakki Motorcu$^{1}$ \qquad Mujdat Cetin$^{1,2,3}$}

\vspace{0.3cm} 

\IEEEauthorblockA{\textit{$^1$Computer Science Department, University of Rochester}}\\
\IEEEauthorblockA{\textit{$^2$Goergen Institute for Data Science \& AIS, University of Rochester}}\\
\IEEEauthorblockA{\textit{$^3$Electrical and Computer Engineering Department, University of Rochester}}

}

\begin{document}
\maketitle


\begin{abstract}
Spatially varying image deblurring remains a fundamentally ill-posed problem, especially when degradations arise from complex mixtures of motion and other forms of blur under significant noise. State-of-the-art learning-based approaches generally fall into two paradigms: model-based deep unrolling methods that enforce physical constraints by modeling the degradations, but often produce over-smoothed, artifact-laden textures, and generative models that achieve superior perceptual quality yet hallucinate details due to weak physical constraints. In this paper, we propose a novel framework that uniquely reconciles these paradigms by taming a powerful generative prior with explicit, dense physical constraints. Rather than oversimplifying the degradation field, we model it as a dense continuum of high-dimensional compressed kernels, ensuring that minute variations in motion and other degradation patterns are captured. We leverage this rich descriptor field to condition a ControlNet architecture, strongly guiding the diffusion sampling process. Extensive experiments demonstrate that our method effectively bridges the gap between physical accuracy and perceptual realism, outperforming state-of-the-art model-based methods as well as generative baselines in challenging, severely blurred scenarios.
\end{abstract}
\section{Introduction}
\label{sec:intro}


Spatially varying image blur remains one of the most challenging problems in image restoration. In real‐world scenarios, degradations can arise from complex mixtures of motion, defocus caused by scene depth, lens aberrations and significant measurement noise. These phenomena combine to produce a degradation field that changes across the image and over time, making the inverse problem severely ill‐posed. Recovering sharp imagery under such variable blur is critical for applications in microscopy, aerial imaging, autonomous sensing and consumer photography, where fine texture and structural fidelity matter greatly.

Approaches to image deblurring differ in how they treat the unknown degradation process. Blind methods attempt to estimate both the latent sharp image and the blur kernel simultaneously, which is more flexible but also severely ill-posed and sensitive to noise. Non-blind methods, by contrast, assume that the degradation model or kernel information is available or pre-estimated, allowing more reliable recovery under controlled or partially calibrated settings. Within these categories, traditional variational approaches rely on analytical priors and explicit regularization, while modern deep learning–based models learn direct mappings from blurry to sharp images using convolutional or transformer architectures. A growing class of hybrid unrolled networks embeds physical models within iterative optimization frameworks, offering a balance between interpretability and data-driven learning. Most recently, generative diffusion models have introduced powerful image priors capable of synthesizing perceptually realistic detail, yet they often lack physical grounding and can hallucinate content inconsistent with the measurements.

In this paper, we propose Physics-Guided ControlNet (PG-ControlNet), a conditional diffusion framework that integrates explicit physical modeling with the generative flexibility of diffusion priors. Our approach operates in the non-blind setting, where spatial information about the degradation process is known or estimated. Instead of simplifying the blur as uniform or region-wise constant, we represent it as a dense field of locally compressed point spread functions, capturing subtle spatial variations caused by motion, defocus, and optical aberrations. This physically grounded descriptor field serves as a conditioning signal for a ControlNet-based diffusion model, steering the sampling process toward solutions that are consistent with the image formation model while maintaining rich perceptual detail. By embedding physical knowledge directly into the diffusion process, PG-ControlNet bridges the divide between physically accurate reconstruction and perceptually realistic image synthesis.

\noindent \textbf{Contributions.}
Our work makes the following contributions:

\begin{itemize}
    \item We introduce a \textit{dense physics-guided blur descriptor} that compactly encodes local kernel variability across the image, enabling fine-grained modeling of spatially varying motion and defocus effects.
    
    \item We design a \textit{conditional diffusion framework} based on ControlNet, where this dense descriptor field modulates the generative process, steering sampling toward physically consistent yet perceptually rich reconstructions.

\end{itemize}
Extensive experiments demonstrate that PG-ControlNet attains state-of-the-art perceptual quality and competitive reconstruction fidelity under challenging spatially varying blur. Ablation studies and qualitative visualizations further confirm that explicit physical conditioning enhances stability and preserves detail across diverse degradation patterns.

To summarize, PG-ControlNet leverages dense, physically grounded conditioning to align the strengths of model-based and generative paradigms. The following sections detail each component of our approach. Section~\ref{sec:related_work} reviews related work. Section~\ref{sec:method} formulates the problem and describes our blur descriptor representation, followed by the conditional diffusion architecture and training strategy. Section~\ref{sec:experiments} reports quantitative and qualitative results together with ablation analyses, and Section~\ref{sec:conclusion} concludes the paper and outlines future directions.




\section{Related Work}
\label{sec:related_work}

Image deblurring methods can be broadly categorized as \textit{blind} or \textit{non-blind}, depending on whether the degradation model is known. Blind approaches aim to jointly estimate the latent sharp image and the blur kernel from a single observation~\cite{levin_blind_2009,fergus2006removing,pan_blind_2016,krishnan2011blind}. Although flexible, this joint estimation is severely ill-posed and highly sensitive to noise and initialization, often requiring strong regularization or handcrafted priors to stabilize the optimization. In contrast, non-blind methods assume that the blur kernel is available through calibration or prior estimation~\cite{Schuler2013AML,hirsch2011fast,fortunato2014fast}. This assumption enables more direct enforcement of data fidelity and allows explicit incorporation of physical knowledge into the reconstruction. When the degradations vary over the spatial field, the deblurring problem becomes even more challenging. Our framework follows the non-blind formulation, leveraging dense degradation priors to guide the restoration process under spatially varying blur.

\noindent \textbf{Deep Learning and Hybrid Unrolling Methods.}
Deep learning has substantially advanced both blind and non-blind image restoration by enabling networks to handle complex, spatially varying degradations without requiring explicit modeling of the underlying blur process. 
CNN- and Transformer-based architectures such as MPRNet \cite{mprnet}, Restormer \cite{restormer}, SwinIR \cite{swinir}, and ESRGAN \cite{esrgan} demonstrate impressive perceptual quality on blur benchmarks. 
More recent architectures including HINet \cite{hinet}, Uformer \cite{uformer}, MIMO-UNet++ \cite{mimounet}, NAFNet \cite{nafnet}, and AdaIR \cite{adair} further refine hierarchical feature aggregation, attention mechanisms, and task adaptability. 
Despite these advances, such purely data-driven models rely on implicit learning of degradation statistics from large paired datasets, which limits their robustness when the test blur distribution deviates from training conditions. 
In practice, they may produce over-smoothed textures in heavily blurred regions and excessive sharpening in low-texture or defocused areas. Recent efforts also leverage semantic priors to improve robustness. SAM-Deblur~\cite{Li2023SAMDeblur} integrates segmentation cues from the Segment Anything Model (SAM) \cite{SAM} to enhance non-uniform deblurring and generalization under spatially varying blur.

Plug-and-play frameworks \cite{plugandplay} and deep unrolling methods \cite{admmnet} have established an effective bridge between traditional optimization-based reconstruction and modern deep learning. By replacing analytical priors with learned denoisers and introducing deep learning based components in the iterative reconstruction frameworks, these methods retain physics-based information while benefiting from learned priors. 
Building on this foundation, DMBSR \cite{dmbsr} unfolded a linearized ADMM scheme into a deep architecture that jointly performs denoising, deblurring, and upsampling under spatially varying blur. These model-based networks are highly effective in incorporating physical constraints into the reconstruction process, leading to improved stability and convergence. However, in severely ill-posed cases where fine textures or perceptual details are lost, their iterative refinement and fixed regularization often produce residual artifacts or incomplete texture recovery. While their structure preserves edge consistency and spatial fidelity, it lacks the generative flexibility required to plausibly reconstruct high-frequency details beyond the physical observation model.

\noindent \textbf{Diffusion-Based Image Restoration.}
While model-based approaches are constrained by fixed priors and limited generative capacity, recent advances in diffusion models have introduced powerful data-driven priors capable of representing complex, high-dimensional image distributions. Denoising Diffusion Probabilistic Models (DDPM) \cite{ddpm} and Latent Diffusion Models (LDM) \cite{ldm} learn the statistical manifold of natural images through iterative denoising, enabling realistic synthesis and restoration. Extensions such as SR3 \cite{sr3}, Palette \cite{palette}, and RePaint \cite{repaint} adapt this principle to conditional generation tasks including super-resolution and inpainting.  Models such as DeblurDiff \cite{kong2025deblurdiff} and ResShift \cite{resshift} specifically address motion and mixed deblurring using learned diffusion priors. These models achieve remarkable perceptual realism and recover fine details often lost in model-based or CNN restorers. However, guiding the diffusion sampling process toward physically consistent reconstructions remains challenging. In the absence of strong local constraints, blind diffusion restorers can hallucinate structures or introduce inconsistencies, particularly under severe or spatially varying degradations.

\noindent \textbf{Conditional and Physics-Guided Diffusion.}
Conditional diffusion models extend the generative framework by incorporating external information to steer the sampling process. Frameworks such as SDEdit \cite{sdedit}, RePaint \cite{repaint}, InstructPix2Pix \cite{instructpix2pix}, and ControlNet \cite{controlnet} demonstrate that spatial or semantic conditioning can produce controllable and locally consistent generations. Training or fine-tuning large diffusion models from scratch, however, is computationally demanding and data-intensive. ControlNet offers a practical alternative by freezing the pretrained diffusion backbone and learning a lightweight adapter that injects conditioning signals, enabling efficient control with limited data. Although such conditional mechanisms have been primarily applied to generation and editing, they provide an appealing foundation for physically informed image restoration.

Score-based generative modeling \cite{songscore} provides a continuous-time formulation of diffusion processes and forms the theoretical basis for many recent physics-aware variants. Recent physics-aware approaches, including IR-Bridge \cite{irbridge}, DPS \cite{dps}, DPS+ \cite{dpsplus}, and DiffPIR \cite{diffpir}, introduce measurement-consistency terms but are typically instantiated with spatially invariant or global degradation operators and do not explicitly model spatially varying blur fields.
In this paper, we adapt the ControlNet paradigm to the restoration domain through a dense, physically grounded descriptor field that encodes spatially varying blur characteristics. This design allows the diffusion model to adapt its generative prior locally according to the degradation pattern, effectively uniting the interpretability of physics-based modeling with the perceptual realism of diffusion-based generation.

\section{Methodology}
\label{sec:method}

\begin{figure}[t]
    \centering
    \includegraphics[width=\linewidth]{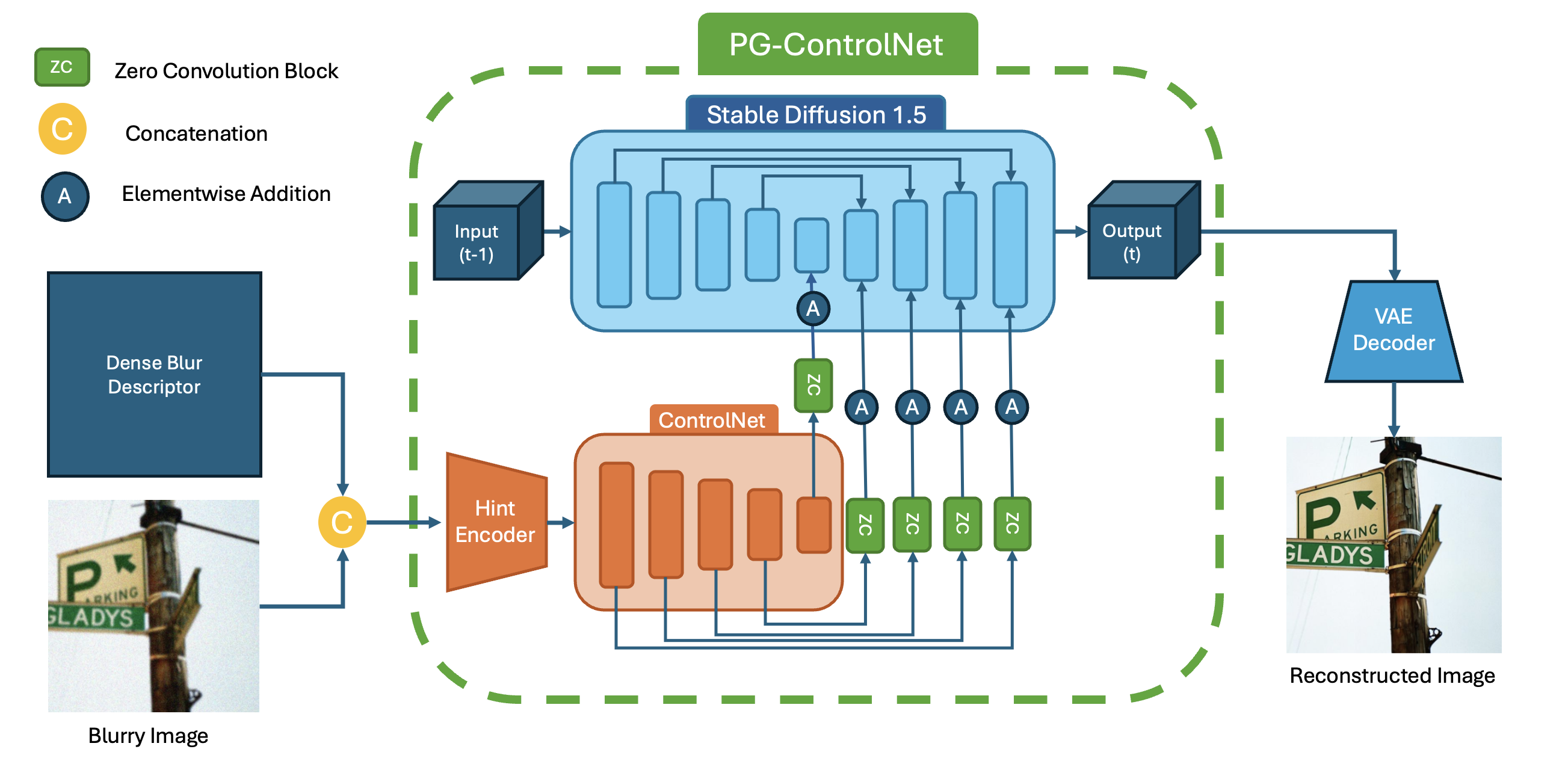}
    \caption{
    Overview of the proposed \textbf{PG-ControlNet} framework. 
    The frozen Stable Diffusion 1.5 backbone and VAE are depicted blue. The text encoder and time inputs are omitted for simplicity. The ControlNet and the \textit{hint encoder} are trained on concatenated inputs of the blurry image $\mathbf{y}$ and its dense blur descriptor field $\mathbf{D}$. 
    }
    \label{fig:architecture}
\end{figure}

Our goal is to recover underlying scenes from spatially varying blurred images by combining explicit physical modeling with the generative expressiveness of diffusion priors. 
To this end, we propose \textit{PG-ControlNet}, a physics-guided conditional diffusion framework that leverages dense degradation descriptors to guide the restoration process. 
Unlike purely data-driven or blind generative methods, PG-ControlNet assumes that a spatially varying blur field is available or estimated, and uses it to provide fine-grained conditioning throughout the diffusion sampling process. 
This section first formalizes the underlying degradation model and training data generation pipeline (\S\ref{sec:forward_model}), then introduces our compact representation of local blur kernels (\S\ref{sec:descriptor_field}), and finally details the proposed physics-guided conditional diffusion architecture and its training procedure (\S\ref{sec:pgcontrolnet}). 
An overview of the complete framework is illustrated in Fig.~\ref{fig:architecture}.
\begin{figure}[t]
    \centering
    \includegraphics[width=\linewidth]{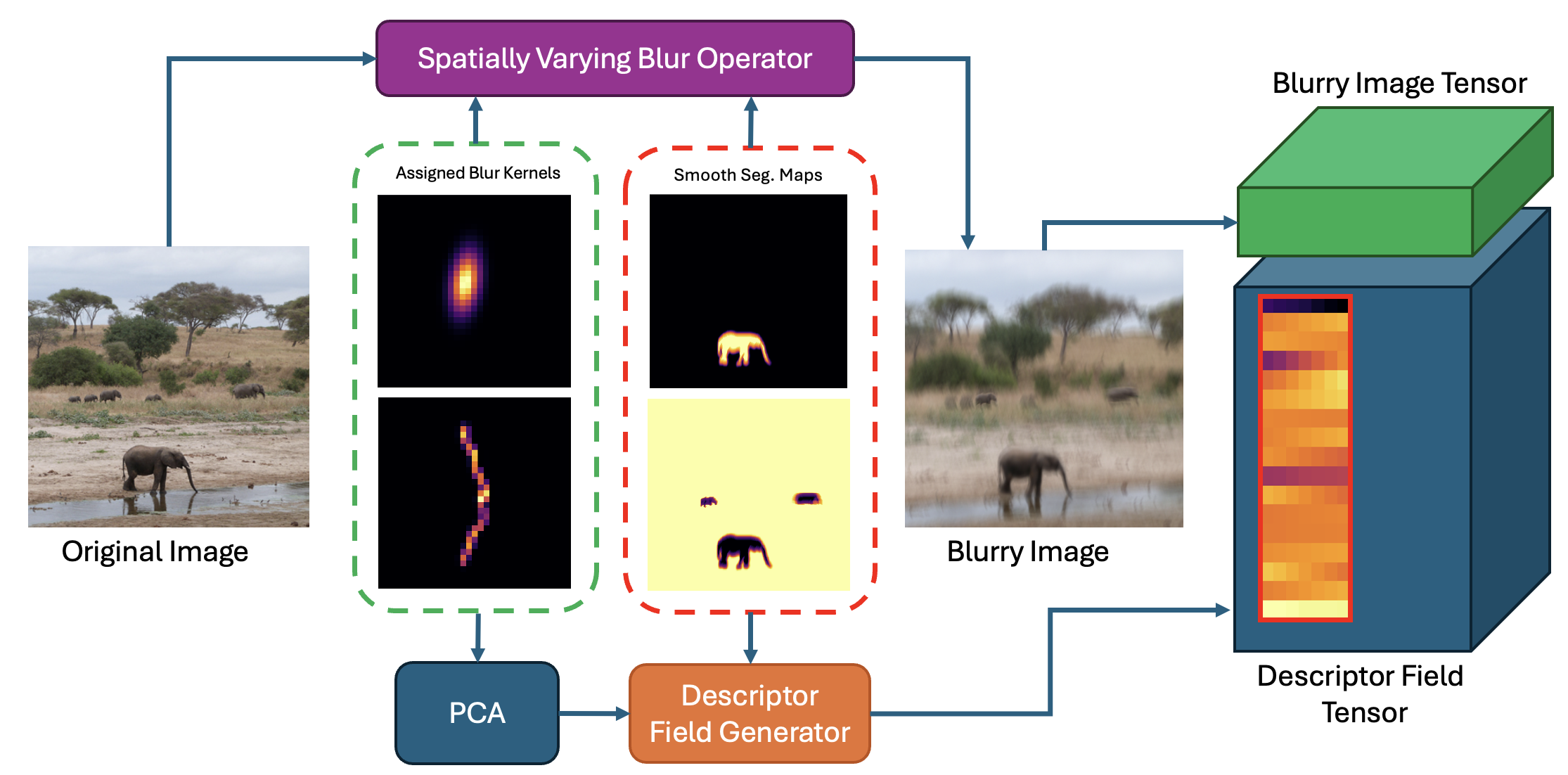}
\caption{
    \textbf{Spatially varying data generation and dense blur descriptor construction pipeline.} 
    (Top) Soft region masks derived from segmentation maps are used to assign region-specific point spread functions (PSFs) $K_i$ to different objects, synthesizing the blurred observation $\mathbf{y}$ according to Eq.~\ref{eq:forward_model}.
    (Bottom) To construct the dense descriptor field $\mathbf{D}$, each $K_i$ is vectorized and PCA-compressed. These embeddings are then spatially distributed by weighting them with the same smooth normalized maps to ensure continuous transitions at region boundaries.
    (Right) The final input to the ControlNet Hint Encoder is a \textbf{concatenated tensor} formed by stacking the blurry image (green block) and the dense descriptor field (blue block) channel-wise.The red frame highlights an \textbf{example stack of PCA vectors on a segmentation map transition region}, illustrating the interpolated conditioning signal resulting from the weighted combination of neighboring kernel embeddings.
}\label{fig:data_descriptor_pipeline}
\end{figure}

\begin{figure*}[t!]
\centering

\includegraphics[width=0.9\linewidth]{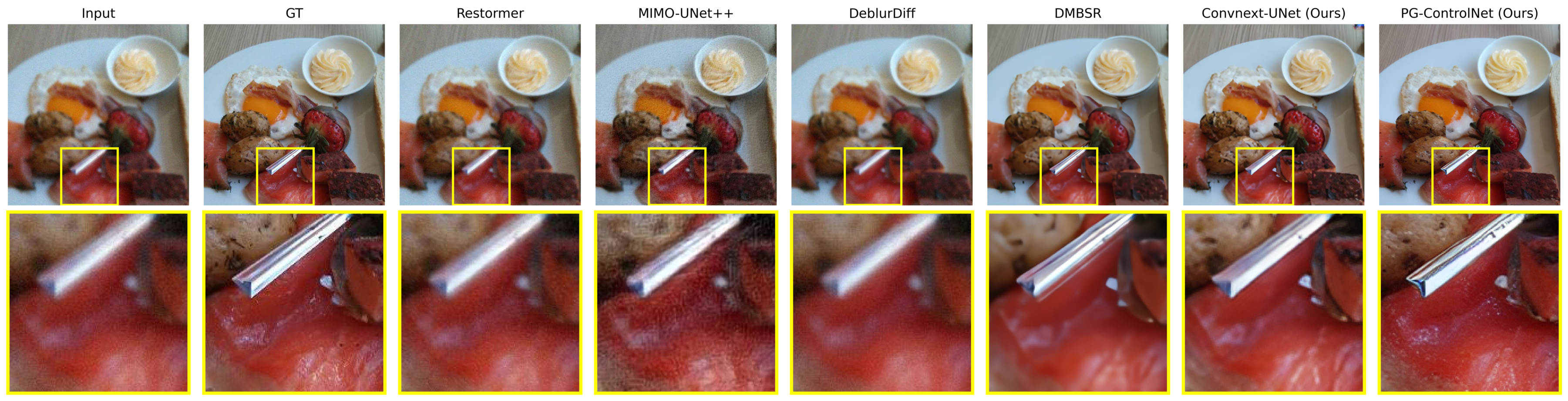}\vspace{-1mm}
\includegraphics[width=0.9\linewidth]{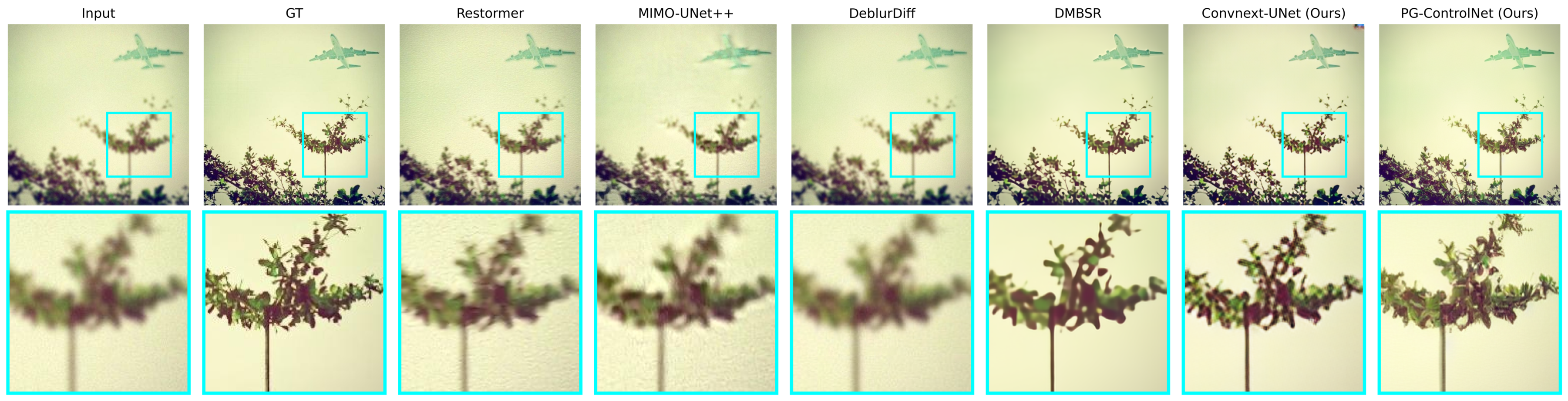}\vspace{-1mm}
\includegraphics[width=0.9\linewidth]{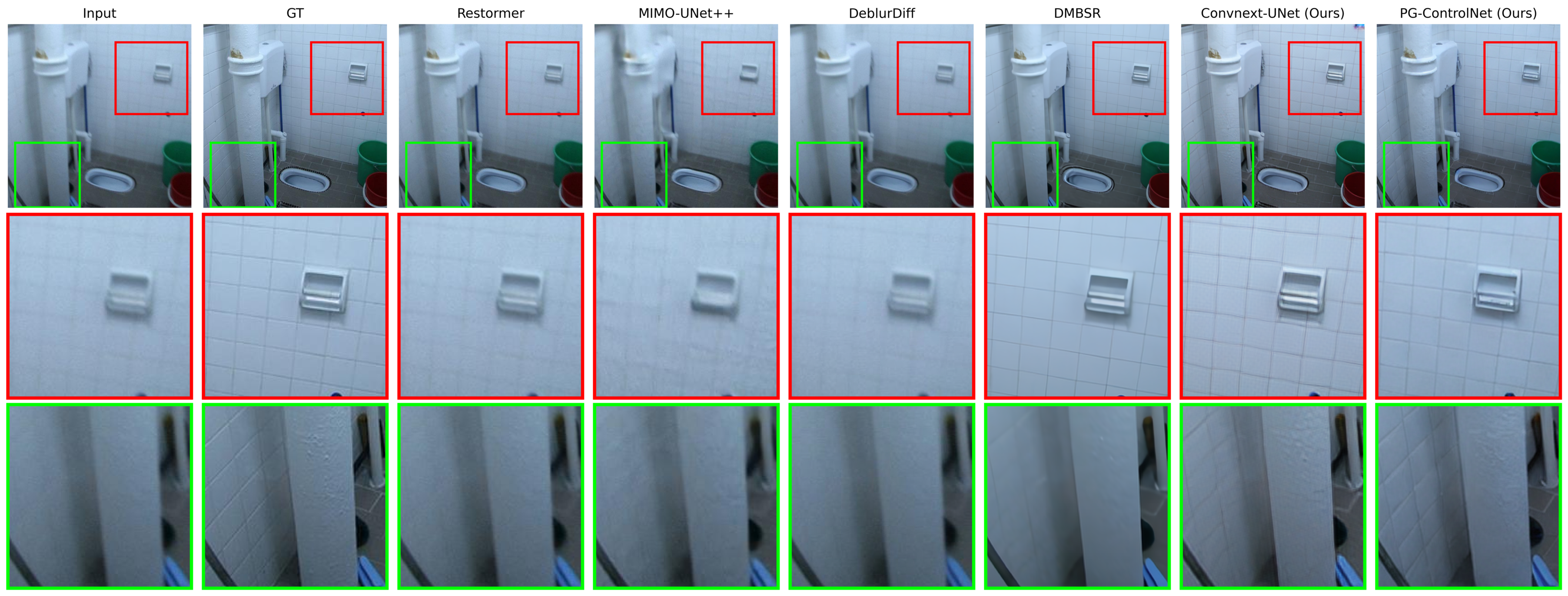}\vspace{-1mm}
\includegraphics[width=0.9\linewidth]{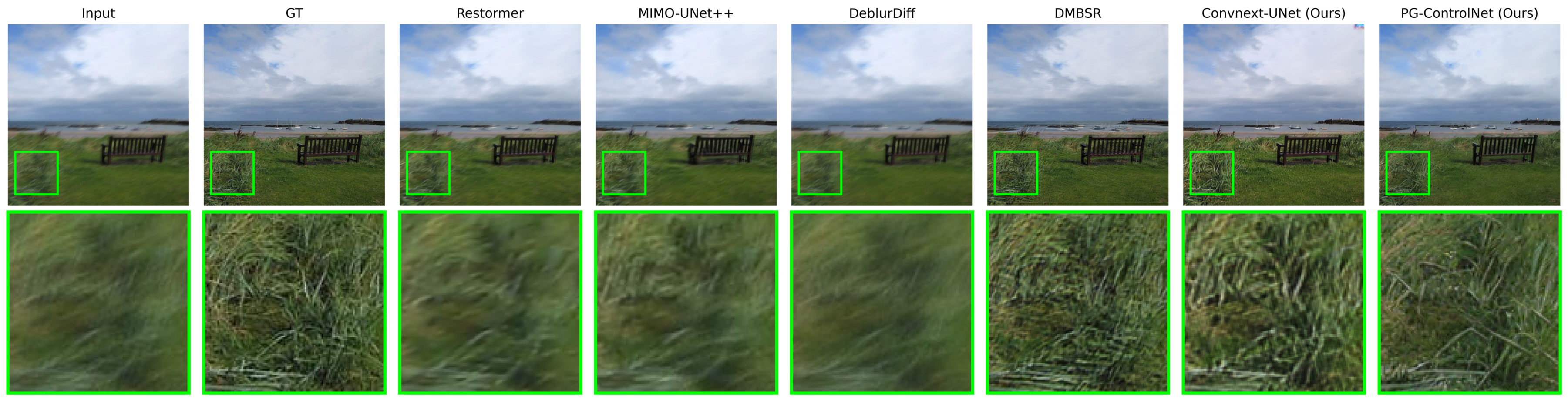}\vspace{-1mm}
\includegraphics[width=0.9\linewidth]{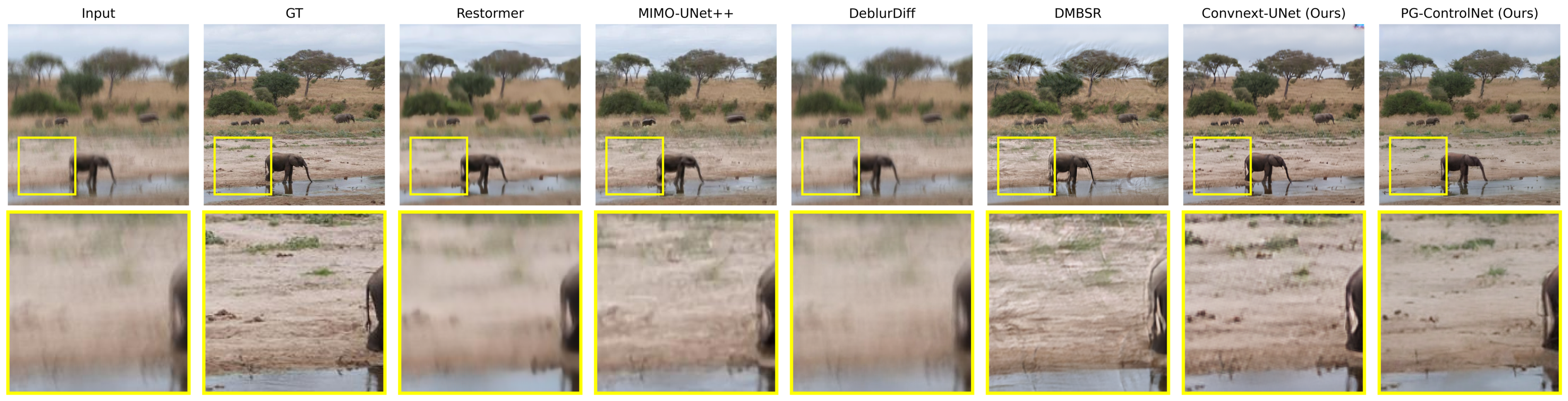}

\caption{
Qualitative comparison on diverse scenes with spatially varying blur. 
Columns show: input, ground truth, Restormer~\cite{restormer}, 
MIMO-UNet++~\cite{mimounet}, DeblurDiff~\cite{kong2025deblurdiff},DMBSR~\cite{dmbsr},  
our non-generative Convnext-UNet, and the proposed PG-ControlNet. 
Zoomed regions are shown beneath each reconstruction.
}
\label{fig:qualitative_comparison}
\end{figure*}
\subsection{Problem Formulation}
\label{sec:forward_model}

We consider the problem of restoring a sharp latent image $\mathbf{x} \in \mathbb{R}^{H \times W \times 3}$ from a spatially varying blurred observation $\mathbf{y} \in \mathbb{R}^{H \times W \times 3}$. 
Following the formulation of O'Leary \textit{et al.} \cite{nagy1998restoring}, the forward degradation model can be expressed as a spatially varying convolution:
\begin{equation}
\mathbf{y} = \sum_{i=1}^{N} M_i \odot (K_i \otimes \mathbf{x}) + \mathbf{n},
\label{eq:forward_model}
\end{equation}

where $M_i$ denotes the spatial weighting mask for region $i$, $K_i \in \mathbb{R}^{k \times k}$ is the corresponding point spread function (PSF), $\otimes$ denotes convolution, $\odot$ denotes element-wise multiplication, and $\mathbf{n}$ represents additive Gaussian noise with random strength. 
To simulate realistic, region-aware degradations, the masks $\{M_i\}$ are derived from semantic segmentation maps of the scene. This allows us to assign distinct blur characteristics (e.g., motion blur to a foreground object and defocus blur to the background) to different semantic regions.
These masks form a smooth partition of unity, ensuring gradual transitions between neighboring blur regions rather than hard cutoffs, similar to the data generation strategy used in DMBSR~\cite{dmbsr}. 
For training data generation, each kernel $K_i$ is randomly sampled from a library of motion~\cite{motion_kernel} and defocus PSFs.
To promote generalization, a small fraction ($1\%$) of regions are left unblurred.
The entire synthesis pipeline is implemented with batched GPU parallelization on Pytorch, enabling efficient generation of large-scale datasets that exhibit realistic spatially varying degradations.

In the non-blind setting considered in this work, the degradation field $H=\{M_i,K_i\}_{i=1}^{N}$ is assumed to be known or estimated beforehand. Given the observation $\mathbf{y}$ and its associated blur operator $H$, the objective is to draw physically consistent samples of the latent clean image $\mathbf{x}$ from the conditional distribution $p_\theta(\mathbf{x}\,|\,\mathbf{y},H)$. Rather than explicitly optimizing a reconstruction objective, we adopt a generative inference approach in which the diffusion process is guided by the known degradation field. This formulation effectively performs posterior sampling under the learned conditional prior, producing solutions that adhere to the physical observation model while remaining perceptually plausible. However, directly conditioning on the full set of local kernels $\{K_i\}$ is computationally expensive and memory-intensive, since each kernel is a high-dimensional matrix and thousands of such kernels exist per image. To address this, we construct a compact \textit{blur descriptor field} that retains the essential local degradation information in a low-dimensional form while remaining spatially aligned with the image. This representation, detailed in Section~\ref{sec:descriptor_field}, serves as a physically interpretable conditioning prior for the diffusion model.

\subsection{Dense Blur Descriptor Representation}
\label{sec:descriptor_field}

Each spatially varying kernel $K_i \in \mathbb{R}^{k \times k}$ encodes information about local degradation characteristics such as motion direction, magnitude, and defocus shape. To build a compact but expressive representation, we apply principal component analysis (PCA) to a large set of training kernels drawn from two sources: the publicly available motion-blur library of \cite{motion_kernel} and an equally sized set of randomly generated Gaussian defocus kernels. The resulting PCA basis captures the dominant modes of variation across both motion and defocus blur, enabling efficient low-dimensional kernel embeddings. The number of components is empirically chosen to preserve fine structural variations while minimizing reconstruction error; a 128-dimensional embedding provides an optimal balance between compactness and fidelity. To form a spatially coherent conditioning map, the local kernel embeddings are organized into a dense descriptor field $\mathbf{D} \in \mathbb{R}^{H \times W \times d}$ that aligns with the image grid. Each pixel location receives a descriptor vector derived from its corresponding region kernel or from a smooth interpolation of neighboring embeddings in transition areas, following the same soft masking strategy used in the forward model. This results in a differentiable, spatially consistent representation that captures gradual variations in the blur field while avoiding discontinuities across boundaries.

\subsection{Physics-Guided Conditional Diffusion Framework}
\label{sec:pgcontrolnet}

Our restoration framework builds upon the Stable Diffusion~1.5 backbone \cite{ldm} and follows the ControlNet \cite{controlnet} conditioning paradigm. The pretrained diffusion model, including its UNet denoiser, VAE encoder–decoder, and text encoder, is kept frozen to preserve its generative prior. A lightweight hint encoder is introduced to inject physics-based information derived from the observed blurred image and its corresponding blur descriptor field. The two inputs  (the RGB blurry image (3 channels) and the PCA descriptor field (128 channels))are concatenated  channel-wise to form a 131-channel tensor, which is processed by the hint encoder to produce multi-scale feature maps. These features are added to the intermediate activations of the frozen UNet at corresponding resolutions, allowing the generative process to be steered by spatially varying physical cues while retaining the expressive prior learned by the diffusion backbone. A fixed text embedding is used throughout training to stabilize sampling and maintain consistency across prompts. An overview of the proposed PG-ControlNet architecture is illustrated in Fig.~\ref{fig:architecture}.

The conditional diffusion process follows the standard formulation of Denoising Diffusion Probabilistic Models (DDPM), where a sample $\mathbf{x}_t$ at noise level $t$ is progressively denoised by a neural estimator $\epsilon_\theta(\mathbf{x}_t, t, c)$ that predicts the additive noise given conditioning information $c$. In our framework, the conditioning variable is defined as $c=(\mathbf{y},\mathbf{D})$, combining the observed blurred image $\mathbf{y}$ and its corresponding blur descriptor field $\mathbf{D}$. The diffusion UNet receives $\mathbf{x}_t$ along with multi-scale feature maps from the hint encoder, enabling the denoiser to exploit both the degraded measurement and the local physical priors during each sampling step. At inference time, we employ the DPMSolverMultistepScheduler with a Karras-style noise schedule for efficient deterministic sampling, which provides faster convergence and improved preservation of fine structures. This configuration allows PG-ControlNet to achieve stable posterior sampling with strong physical consistency while maintaining the perceptual quality of the pretrained diffusion backbone.

\noindent\textbf{Training Objective.}
During training, only the hint encoder parameters are optimized while the diffusion backbone remains frozen. 
The model is trained using the standard noise-prediction objective of diffusion models, 
where the network learns to estimate the additive Gaussian noise given the conditioning inputs $c=(\mathbf{y},\mathbf{D})$. 
A cosine learning rate schedule with an initial rate of $5\times10^{-5}$ and gradient accumulation of 2 steps yields an effective batch size of 42. 
Training is conducted for approximately six days on a single Nvidia A100 GPU with diffusion forward process with 1000 timesteps.
Training pairs $(\mathbf{x},\mathbf{y})$ are synthesized following the forward model described in Section~\ref{sec:forward_model}, using motion and defocus kernels and Gaussian noise with random strength. 
Freezing the pretrained backbone ensures that the generative prior is preserved, while the hint encoder and ControlNet learns to translate spatially varying blur information into control signals that guide the denoising process.

\section{Experiments}
\label{sec:experiments}

\subsection{Dataset and Evaluation Setup}

All experiments are conducted on images derived from the COCO~2017 \cite{lin2014microsoft} dataset. 
Training pairs $(\mathbf{x},\mathbf{y})$ are generated using the spatially varying forward model introduced in Section~\ref{sec:forward_model}. 
Images smaller than $512\times512$ are upsampled to match the native resolution of the Stable Diffusion \cite{ldm} backbone, ensuring consistent spatial coverage and controlled degradation. 
For each image, region masks are assigned randomly and convolved with motion or defocus kernels sampled from the kernel library \cite{motion_kernel} , followed by additive Gaussian noise with random strength uniformly sampled in the standard deviation range of 0.05–0.25. 
Most kernel instances appear during training, while a  subset of unseen motion kernels is reserved for testing to assess generalization. 
All models are trained and evaluated on $512\times512$ RGB images using identical degradation conditions.

\subsection{Baseline Models}

We compare PG-ControlNet with a diverse set of model-based, deep learning, and generative restoration approaches.  
DMBSR~\cite{dmbsr} represents the physics-guided optimization baseline that explicitly models spatially varying priors through unrolled ADMM iterations and serves as a reference for physical accuracy.  
Restormer~\cite{restormer} and MIMO-UNet++~\cite{mimounet} are strong supervised architectures widely used for non-uniform deblurring; both are initialized with official pretrained weights and lightly fine-tuned on our dataset for fair comparison.  
DeblurDiff~\cite{kong2025deblurdiff} serves as a diffusion-based baseline that models degradations implicitly, often producing visually pleasing but less measurement-consistent reconstructions.  
All methods are trained or fine-tuned under identical degradation settings and evaluated using the same data splits.

While our proposed framework involves combining dense degradation descriptors with a generative model, one might be curious about the possibility of pairing our proposed dense descriptor field with a non-generative network as well. To address that, we implement a Convnext-v2 \cite{Convnext} inspired UNet \cite{ronneberger2015u}  trained with the same spatially varying blur priors to replace the diffusion model in PG-ControlNet. We present the results of that Convnext-UNet approach as one of the baselines for comparison, although one might also view that as a non-generative variant of our proposed framework.

\subsection{Qualitative Results}
\label{sec:qualitative_results}

Figure~\ref{fig:qualitative_comparison} shows a visual comparison across representative test scenes with diverse spatially varying blur patterns. Overall, PG-ControlNet delivers perceptually sharp and physically consistent reconstructions with fewer artifacts than competing approaches. Compared to DMBSR~\cite{dmbsr} and the non-generative Convnext-UNet, our diffusion-guided model recovers finer textures and more coherent local structures while avoiding the repetitive or oversharpened artifacts commonly observed in other restorers.

In the first row (breakfast example), which highlights the challenge of recovering mixed textures, DMBSR~\cite{dmbsr} exhibits ringing artifacts around the metallic utensil, and DeblurDiff~\cite{kong2025deblurdiff} hallucinates inconsistent smooth textures. While our non-generative variant recovers the utensil's shape, it renders the utensil and fish textures inaccurately. PG-ControlNet achieves the best perceptual balance, faithfully reconstructing the specular highlights of the metal and the organic texture of the food.

In the second row, which contains fine foliage against a flat sky, PG-ControlNet reconstructs leaf details and sky smoothness faithfully. Restormer~\cite{restormer} introduces artificial textures in the sky and fails to recover the plant structure, while MIMO-UNet++~\cite{mimounet} performs poorly overall. While the non-generative Convnext-UNet and DMBSR produce sharp results, the output appears unnatural and ``cartoonish''. Our method restores the fine foliage and the smooth background while maintaining a natural, photographic appearance.

The third row (bathroom scene) demonstrates robustness to geometric structures. Purely data-driven methods like Restormer~\cite{restormer} and MIMO-UNet++~\cite{mimounet} often wipe out the high-frequency grid lines of the tiles. Both DMBSR~\cite{dmbsr} and PG-ControlNet successfully recover these straight lines; however, a closer look at the green crop reveals that only PG-ControlNet succeeds in reconstructing the tile edges \textit{behind} the pipe and the texture on the pipe, which are lost in all other methods.

The fourth row (grass example) highlights the difference between perceptual realism and over-sharpening. The non-generative Convnext-UNet recovers the main texture content but over-sharpens it, resulting in an unrealistic appearance. In contrast, PG-ControlNet retains the salient features of the grass and fills the ill-posed regions with plausible grass blades, creating a more natural texture.

The final row depicts an extreme case combining one unseen motion blur on the elephant with strong background defocus. Blind diffusion models fail to reconstruct coherent structures, and DMBSR~\cite{dmbsr} introduces ringing artifacts. Notably, the non-generative variant introduces unnatural ``cross-hatch'' patterns---likely an adversarial artifact of minimizing the LPIPS loss. PG-ControlNet avoids such artifacts, restoring both the global background and object details, with minor oversmoothing but superior realism.

Across all examples, PG-ControlNet exhibits strong generalization to unseen degradations, maintaining spatial continuity and perceptual realism while effectively reducing boundary and ringing artifacts.

\subsection{Quantitative Results}

Table~\ref{tab:quantitative_results} summarizes the quantitative comparison across all evaluated methods. 
The non-generative Convnext-UNet achieves the highest PSNR and SSIM values, which is expected given that it is trained directly with distortion-oriented objectives. 
However, this improvement again comes at the expense of perceptual fidelity, as the model tends to introduce over-sharpened and unnatural high-frequency details. 
In contrast, PG-ControlNet obtains the best perceptual quality, achieving the lowest LPIPS score (0.1127), the best FID (44.30), and the highest FSIM (0.9479), while maintaining competitive SSIM (0.7729). 
Compared to DMBSR~\cite{dmbsr}, our framework improves perceptual similarity by more than 40\% in LPIPS while offering comparable reconstruction fidelity. 

Diffusion-based baselines such as DeblurDiff~\cite{kong2025deblurdiff} yield visually appealing but less consistent results, with higher LPIPS and noticeable structural deviations. 
Similarly, non-generative networks like Restormer~\cite{restormer} and MIMO-UNet++~\cite{mimounet} perform reliably on moderate blur but degrade under strong spatial variation, where their implicitly learned priors fail to generalize. The physics-free ControlNet variant further highlights this effect: removing the dense descriptor field leads to noticeably worse perceptual scores and less stable reconstructions, underscoring the importance of explicit physical conditioning.
For completeness, we also report the performance averaged over 30 sampling seeds.  
PG-ControlNet remains stable under sampling variability and continues to outperform the physics-free ControlNet baseline across all perceptual metrics in the averaged case as well. 
Overall, PG-ControlNet successfully balances physical consistency and perceptual realism, outperforming both model-based and blind methods under identical degradation conditions.

\begin{table}[t]
\centering
\resizebox{\linewidth}{!}{
\begin{tabular}{lccccc}
\toprule
\textbf{Method} & \textbf{PSNR}$\uparrow$ & \textbf{SSIM}$\uparrow$ & \textbf{LPIPS}$\downarrow$ & \textbf{FID}$\downarrow$ & \textbf{FSIM}$\uparrow$ \\
\midrule
DMBSR~\cite{dmbsr}                     & 23.98 & 0.7020 & 0.2009 & 57.20 & 0.8727 \\
Restormer~\cite{restormer}             & 23.46 & 0.6719 & 0.3282 & 87.29 & 0.8796 \\
MIMO-UNet++~\cite{mimounet}            & 23.13 & 0.6518 & 0.3148 & 98.95 & 0.8749 \\
DeblurDiff~\cite{kong2025deblurdiff}   & 23.09 & 0.6386 & 0.2986 & 93.50 & 0.8625 \\
\midrule
Convnext-UNet (non-generative variant of PG-ControlNet)                   & \textbf{25.96} & \textbf{0.8138} & 0.1543 & 54.70 & 0.9284 \\
\midrule
\rowcolor{gray!10}
PG-ControlNet (ours, best LPIPS seed)  & 24.59 & 0.7729 & \textbf{0.1127} & \textbf{44.30} & \textbf{0.9479} \\
PG-ControlNet (ours, best SSIM seed)   & 25.39 & 0.7941 & 0.1726 & 50.28 & 0.9344 \\
PG-ControlNet (ours, avg 30 seeds)     & 24.72 & 0.7811 & 0.1519 & N/A$^\dagger$ & 0.9451 \\
ControlNet (w/o physics, avg 30 seeds) & 23.52 & 0.6992 & 0.1926 & N/A$^\dagger$ & 0.9071 \\
\bottomrule
\end{tabular}
}
\caption{
Quantitative comparison on the COCO-2017~\cite{lin2014microsoft} validation subset (512$\times$512). 
PG-ControlNet achieves the best perceptual quality (lowest LPIPS and FID) and the highest FSIM, indicating superior texture and structural preservation while maintaining competitive fidelity. 
For completeness, we report the best-performing seeds for LPIPS and SSIM, as well as the average performance over 30 independent sampling seeds.
$^\dagger$FID is omitted for the averaged result, as multiple reconstructions per image (30 seeds) would distort the distributional comparison underlying FID.
}
\label{tab:quantitative_results}
\end{table}




\subsection{Ablation and Analysis}
\label{sec:ablation}

\begin{figure}[t]
\centering

\begin{subfigure}{\linewidth}
\centering
\includegraphics[width=\linewidth]{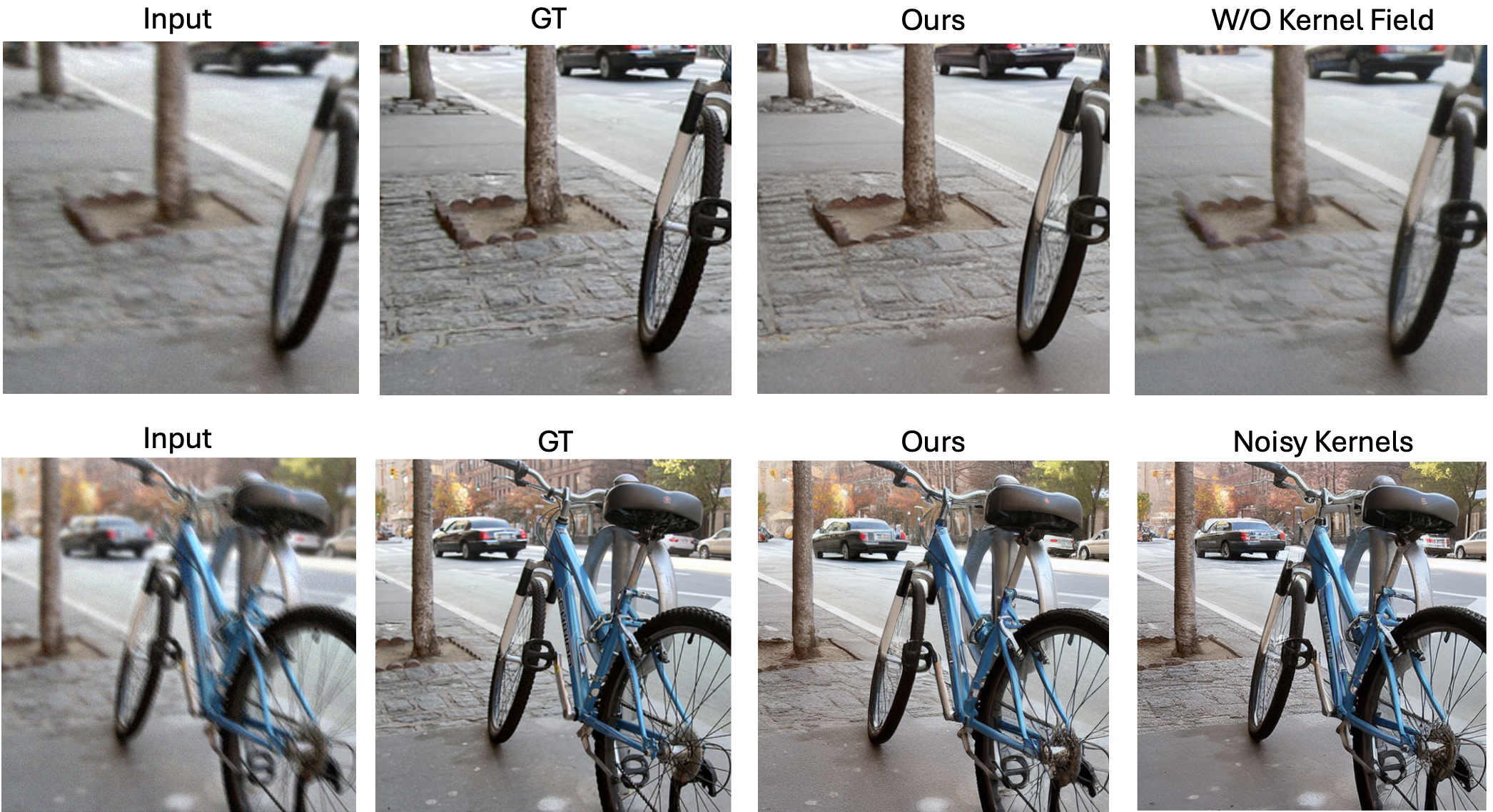}
\end{subfigure}

\vspace{1mm}
\begin{subfigure}{\linewidth}
\centering
\includegraphics[width=\linewidth]{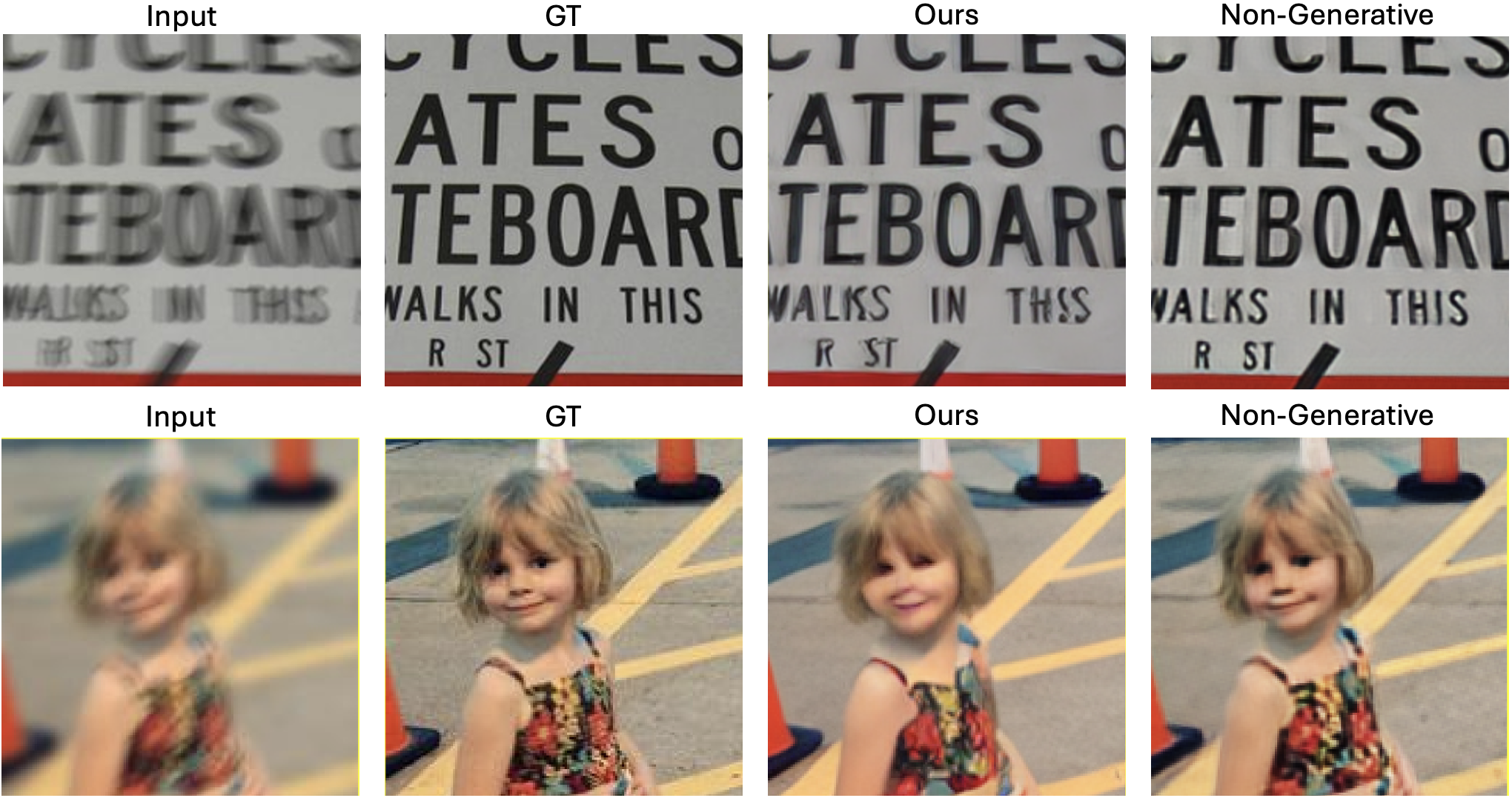}
\end{subfigure}

\caption{
Ablation and limitation analysis.  
Row 1: Effect of removing the kernel field.  
Row 2: Robustness under noisy kernels.  
Rows 3-4: Limitations inherited from the SD-1.5 backbone on text and faces.
}
\label{fig:ablation}
\end{figure}

We first assess the role of the dense descriptor field by comparing PG-ControlNet with a separately trained blind variant conditioned only on the blurry input. This baseline adopts the same architecture and training setup but lacks kernel-field information, effectively turning the model into a purely image-driven ControlNet. As shown in Row~1 of Figure~\ref{fig:ablation}, the blind variant struggles to resolve severe spatially varying blur, leading to inconsistent reconstructions that fluctuate across seeds and scenes—sometimes under-correcting and sometimes over-sharpening. Quantitatively, removing the descriptor field reduces average SSIM from 0.7811 to 0.6992 and increases LPIPS by roughly 21\%, indicating a clear drop in both structural and perceptual fidelity. PG-ControlNet, by contrast, leverages the kernel field to resolve local ambiguity in the degradation pattern, producing sharper and more coherent reconstructions.

We next test the robustness of PG-ControlNet to imperfect degradation priors by perturbing all kernels in the descriptor field with additive Gaussian noise of standard deviation 0.05 after normalization to the [0,1] range (corresponding to an approximate SNR of 18 dB). Despite this corruption, PG-ControlNet remains stable and continues to produce coherent reconstructions, with only minor losses in fine detail. As illustrated in Row~2 of Figure~\ref{fig:ablation}, the model still recovers the dominant spatial blur structure even when the descriptor field is noticeably degraded. This indicates that PG-ControlNet does not rely on highly accurate kernel estimates: coarse spatial cues from the descriptor field are often sufficient to steer the diffusion process toward plausible solutions under moderate prior noise.

We also compare PG-ControlNet with its non-generative Convnext-UNet counterpart trained under identical spatially varying blur priors. Although the Convnext-UNet attains higher SSIM, its combination of MSE, SSIM, and LPIPS training losses often drives the model toward overly sharp or repetitive micro-textures that do not correspond to the true scene content. These artifacts can inflate distortion-oriented metrics while degrading perceptual realism. PG-ControlNet, in contrast, produces more natural high-frequency details and maintains global coherence, reflecting the advantage of diffusion-based sampling guided by the kernel field over deterministic regression.

Despite these strengths, PG-ControlNet still inherits certain limitations of the Stable Diffusion~1.5 backbone \cite{ldm}. While the dense kernel field improves reconstruction quality even in challenging regions, some semantic details—particularly in text and faces—remain difficult to recover. The examples in Figure~\ref{fig:ablation} illustrate this behavior: PG-ControlNet produces coherent global structure and reduces blur effectively, yet some fine details are not fully reconstructed, reflecting known weaknesses of the underlying generative prior. These are not specific to our conditioning mechanism, and future work will explore more capable transformer-based latent diffusion models or task-adapted backbones to further improve semantic fidelity.

\noindent\textbf{Overall Findings.}
The ablation results collectively demonstrate that the dense descriptor field provides interpretable physical guidance, while diffusion-based sampling contributes perceptual diversity and robustness. 
Together, these components provide a step toward bridging the gap between physically constrained and perceptually driven image restoration.

\section{Conclusion and Discussion}
\label{sec:conclusion}

We have presented  PG-ControlNet, a physics-guided conditional diffusion framework for spatially varying image deblurring. By embedding a dense descriptor field that encodes local blur characteristics, our approach reconciles physically grounded modeling with the expressive power of generative diffusion priors. Unlike traditional unrolled or supervised networks, PG-ControlNet achieves a balanced trade-off between measurement fidelity and perceptual realism, producing spatially coherent reconstructions even under complex, nonuniform degradations.

Extensive experiments demonstrate that PG-ControlNet outperforms state-of-the-art model-based, supervised, and diffusion-based baselines in perceptual metrics while maintaining competitive fidelity scores. Ablation studies confirm that the dense conditioning field and the frozen diffusion backbone play complementary roles—providing interpretability and stability without sacrificing generative diversity. The resulting framework offers a principled yet flexible approach for image restoration tasks where spatially varying degradations are either known or estimable.

Our current implementation builds upon Stable Diffusion~1.5, which limits performance on fine semantic structures such as human faces and text due to its pretrained feature distribution. Future work will explore stronger latent priors, adaptive noise schedules, and task-specific backbones to further enhance reconstruction fidelity. Beyond natural imagery, the proposed formulation holds promise for controlled acquisition systems such as microscopy, aerial imaging, and depth-aware photography, where degradation priors can be accurately modeled or measured. We believe that physics-guided generative conditioning opens new directions for interpretable, data-efficient image restoration.

\section*{Acknowledgments}
This work was supported in part by the Arnold and Mabel Beckman Foundation under a grant supporting physics-guided imaging and reconstruction research.

\section*{References}
{
    \small
    \bibliographystyle{IEEEtran}
    \bibliography{main} 
}

\end{document}